\documentclass[a4paper]{article}
\usepackage{INTERSPEECH2015}
\usepackage{epsfig}
\usepackage{algorithm,algorithmic}
\usepackage{amsmath,graphicx,paralist}

\title{The IBM 2016 English Conversational Telephone Speech Recognition System}
\name {George Saon, Tom Sercu, Steven Rennie and Hong-Kwang J. Kuo}
\address{IBM T. J. Watson Research Center, Yorktown Heights, NY, 10598\\
\tt{gsaon@us.ibm.com}}

\ninept
\copyrightnotice{\copyright\ IEEE 2004}
\toappear{Submitted to Interspeech 2016, please do not redistribute}

\begin{document}
\maketitle

\begin{abstract} 

  We describe a collection of acoustic and language modeling
  techniques that lowered the word error rate of our English
  conversational telephone LVCSR system to a record 6.6\% on the
  Switchboard subset of the Hub5 2000 evaluation testset. On the
  acoustic side, we use a score fusion of three strong models:
  recurrent nets with maxout activations, very deep convolutional nets
  with 3x3 kernels, and bidirectional long short-term memory nets which
  operate on FMLLR and i-vector features. On the language modeling side, we
  use an updated model ``M'' and hierarchical neural network LMs.

\end{abstract}


\section{Introduction}

The landscape of neural network acoustic modeling is rapidly evolving.
Spurred by the success of deep feed-forward neural nets for LVCSR
in~\cite{seide11} and inspired by other research areas like image
classification and natural language processing, many speech groups
have looked at more sophisticated architectures such as deep
convolutional nets~\cite{abdelhamid13, sainath13}, deep recurrent
nets~\cite{saon14}, time-delay neural nets~\cite{peddinti15}, and
long-short term memory nets~\cite{hannun14, sak15, miao15, mohamed15}. The trend is
to remove a lot of the complexity and human knowledge that was
necessary in the past to build good ASR systems (e.g. speaker
adaptation, phonetic context modeling, discriminative feature
processing, etc.) and to replace them with a powerful neural network
architecture that can be trained agnostically on a lot of
data. With the advent of numerous neural network toolkits which can
implement these sophisticated models out-of-the-box and powerful
hardware based on GPUs, the barrier of entry for building high
performing ASR systems has been lowered considerably. First case in
point: front-end processing has been simplified considerably with the
use of CNNs which treat the log-mel spectral representation as an
image and don't require extra processing steps such as PLP cepstra,
LDA, FMLLR, fMPE transforms, etc. Second case in point: end-to-end ASR
systems such as~\cite{hannun14,miao15,sak15} bypass the need of having
phonetic context decision trees and HMMs altogether and directly map
the sequence of acoustic features to a sequence of characters or
context independent phones. Third case in point: training algorithms
such as connectionist temporal classification~\cite{graves13} don't require an initial
alignment of the training data which is typically done with a
GMM-based baseline model.

The above points beg the question whether, in this age of readily
available NN toolkits, speech recognition expertise is still necessary
or whether one can simply point a neural net to the audio and
transcripts, let it train, and obtain a good acoustic model. While it
is true that, as the amount of training data increases, the need for
human ASR expertise is lessened, at the moment the performance of
end-to-end systems ultimately remains inferior to that of more
traditional, i.e.  HMM and decision tree-based, approaches.  Since the
goal of this work is to obtain the lowest possible WER on the
Switchboard dataset regardless of other practical considerations such
as speed and/or simplicity, we have focused on the latter approaches.

The paper is organized as follows: in section~\ref{sys} we discuss
acoustic and language modeling improvements and in
section~\ref{conclusion} we summarize our findings.

\section{System improvements}
\label{sys}

In this section we describe three different acoustic models that were
trained on 2000 hours of English conversational telephone speech:
recurrent nets with maxout activations and annealed dropout, very deep
convolutional nets with 3$\times$3 kernels, and bidirectional
long short-term memory nets operating on FMLLR and i-vector features. All
models are used in a hybrid HMM decoding scenario by subtracting the
logarithm of the HMM state priors from the log of the softmax output
scores.

The training and test data,
frontend processing, speaker adaptation are identical to~\cite{saon15}
and their description will be omitted. At the end of the section, we
also provide an update on our vocabulary and language modeling
experiments.

\subsection{Recurrent nets with maxout activations}

\begin{figure}[htpb!]
\centerline{\includegraphics[scale=0.7]{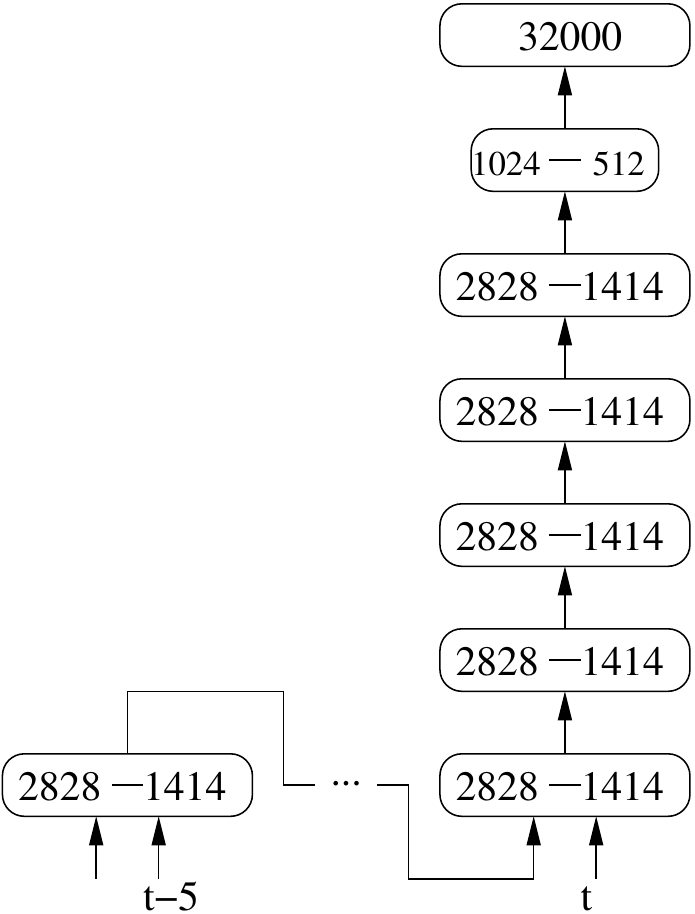}}
\caption{\label{rnn} Unfolded maxout RNN architecture (right arrows in the boxes denote the maxout operation).}
\end{figure}

We remind the reader that maxout nets~\cite{goodfellow13} generalize
ReLu units by employing non-linearities of the form $s_i = \max_{j\in
  C(i)} \left\{{\bf w}^T_{j}{\bf x} +b_j\right\}$ where the subsets of
neurons $C(i)$ are typically disjoint. In~\cite{saon15} we have shown
that maxout DNNs and CNNs trained with annealed dropout outperform
their sigmoid-based counterparts on both 300 hours and 2000 hours
training regimes. What was missing there was a comparison between
maxout and sigmoid for unfolded RNNs~\cite{saon14}. The architecture
of the maxout RNNs comprises one recurrent layer with 2828 units
projected to 1414 units via non-overlapping $2\rightarrow 1$ maxout
operations. This layer is followed by 4 non-recurrent layers with 2828
units (also projected to 1414) followed by a bottleneck with
1024$\rightarrow$512 units and an output layer with 32000 neurons
corresponding to as many context-dependent HMM states.  The number of
neurons for the maxout layers have been chosen such that the weight
matrices have roughly the same number of parameters as the baseline
sigmoid network which has 2048 units per hidden layer. The recurrent
layer is unfolded backwards in time for 6 time steps $t-5\ldots t$ and
has 340-dimensional inputs consisting of 6 spliced right context
40-dimensional FMLLR frames ($t\ldots t+5$) to which we append a
100-dimensional speaker-based ivector. The unfolded maxout RNN
architecture is depicted in Figure~\ref{rnn}.

The network is trained one hidden layer at a
time with discriminative pretraining followed by 12 epochs of SGD CE
training on randomized minibatches of 250 samples. The model is refined with
Hessian-free sequence discriminative training~\cite{bedk12} using the state-based
MBR criterion for 10 iterations.

\begin{table}[htpb!]
\begin{center}
\begin{tabular}{|l|c|c|} \hline
     Model   & WER SWB & WER CH      \\ \hline
RNN sigmoid (CE) & 10.8 & 16.9        \\ \hline
RNN maxout  (CE) & 10.4 & 16.2 \\ \hline
RNN maxout  (ST) & 9.3  & 15.4 \\ \hline
\end{tabular}
\end{center}
\caption{\label{maxout}
Word error rates for sigmoid vs. Maxout RNNs trained with
annealed dropout on Hub5'00 after cross-entropy training (and sequence training for Maxout).}
\end{table}

In Table~\ref{maxout} we report the error rates for sigmoid and maxout
RNNs on the Switchboard and CallHome subsets of Hub5'00. The decodings
are done with a small vocabulary of 30K words and a small 4-gram
language model with 4M n-grams.  Note that the sigmoid RNNs have
better error rates than what was reported in~\cite{saon15} because
they have been retrained after the data has been realigned with the
best joint RNN/CNN model. We observe that the maxout RNNs are
consistently better and that, by themselves, they achieve a similar
WER as our previous best model which was the joint RNN/CNN with
sigmoid activations.

\subsection{Very deep convolutional networks}
Very deep CNNs with small $3\times3$ kernels have recently been shown to achieve
very strong performance as acoustic models in hybrid NN-HMM speech recognition systems.
Results were provided after cross-entropy training on the 300 hours switchboard-1 dataset in \cite{sercu2015very},
and results from sequence training on both switchboard-1 and the 2000 hours switchboard+Fisher dataset are in \cite{sercu2016advances}.

The very deep convolutional networks are inspired by the ``VGG Net'' architecture 
introduced in \cite{simonyan2014very} for the 2014 ImageNet classification challenge,
with the central idea to replace large convolutional kernels by small $3\times3$ kernels.
By stacking many of these convolutional layers with ReLU nonlinearities before pooling layers,
the same receptive field is created with less parameters and more nonlinearity.

\begin{figure*}[ht]
    \centering
    \includegraphics[width=0.7\linewidth]{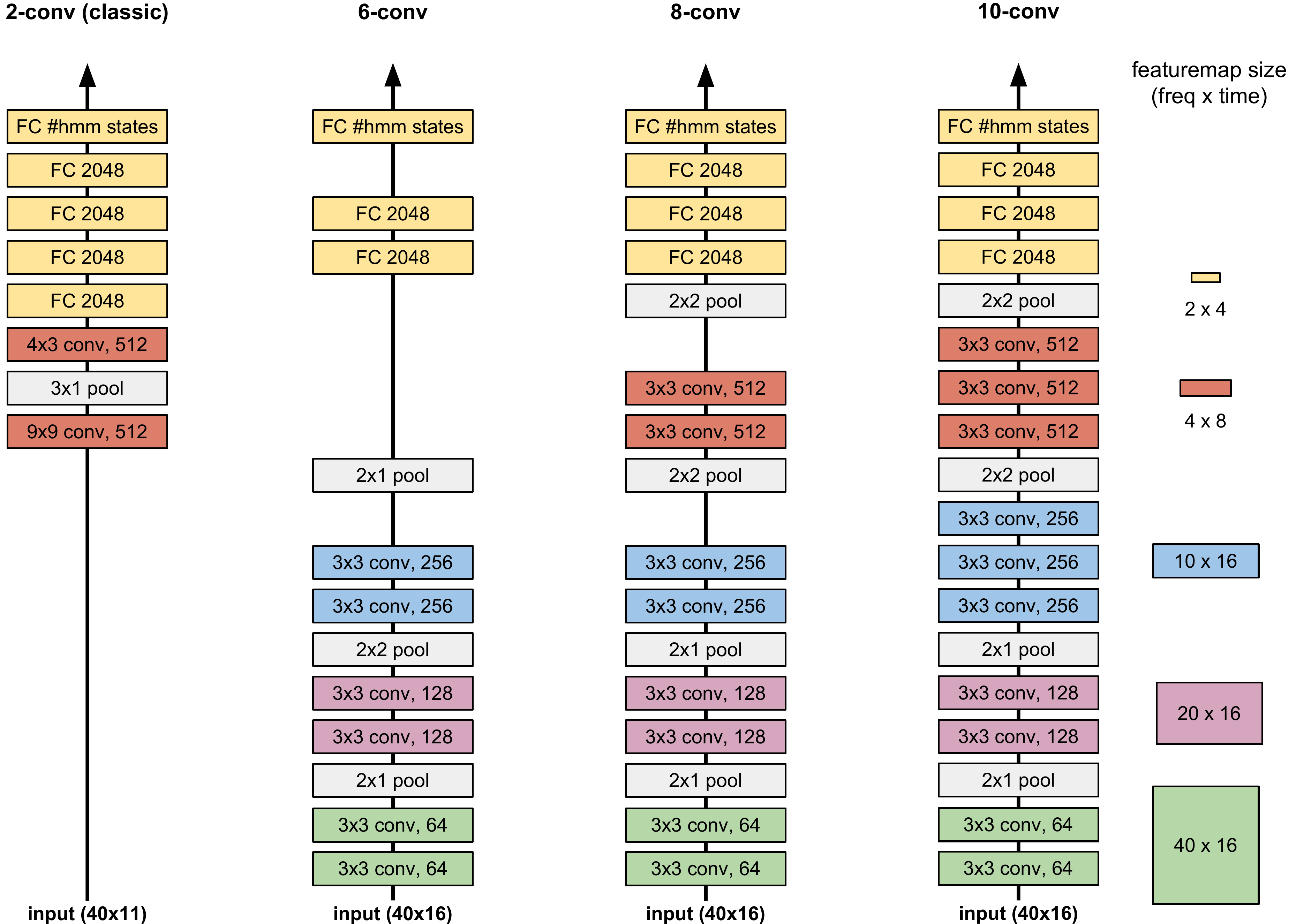}
    \caption{The design of the VGG nets: (1) classical CNN, (2-4) very deep CNNs from \cite{sercu2015very} 
    with 6, 8 and 10 convolutional layers respectively. The deepest CNN (10-conv) obtains best performance.
    This figure corresponds to \cite{sercu2015very} Table 1. }
    \label{fig:vggnets}
\end{figure*}

Figure \ref{fig:vggnets} shows the design of the networks.
Note that as we go deeper in the network, the time and frequency resolution is reduced
through pooling only, while the convolutions are zero-padded as to not reduce the size of the feature maps.
We increase the number of feature maps gradually from 64 to 512 (indicated by the different colors).
We pool right before the layer that increases the number of feature maps.
Note that the indication of feature map size on the right only applies to the rightmost 2 designs.
In contrast, the classical CNN architecture has only two layers, goes to 512 feature maps directly,
and uses a large $9\times9$ kernel on the first layer.
Our 10-layer CNN has about the same number of parameters as the classical CNN,
converges in 5 times fewer epochs, but is computationally more expensive.

Results for 3 variations of the 10-layer CNN are in table \ref{tab:abc}.
For model combination, we use the version with pooling, which is the exact same model without 
modifications from the original paper \cite{sercu2015very}.

\begin{table}[ht]
\centering
\begin{tabular}{| l | c | c | c | c |}\hline
    CNN model & \multicolumn{2}{|c|}{SWB (300h)} & \multicolumn{2}{|c|}{SWB (2000h)} \\ \cline{2-5}
                                         & CE   & ST   & CE   & ST   \\ \hline
    Classic sigmoid \cite{soltau2014joint}   & 13.2 & 11.8 & --    & --    \\
    Classic maxout \cite{saon15} & 12.6 & 11.2 & 11.7* & 9.9* \\  
    \hline
    (a) 10-conv Pool                             & 11.8 & 10.5 & 10.2 & \textbf{9.4}   \\
    (b) 10-conv No pool                          & 11.5 & 10.9 & 10.7 & 9.7  \\ 
    (c) 10-conv No pool, no pad                  & 11.9 & 10.8 & 10.8 & 9.7  \\ \hline
\end{tabular}
\caption{\label{tab:abc} WER on the SWB part of the Hub5'00 testset,
    for 3 variants of the 10-convolutional-layer CNN: with pooling in time (a), without pooling in time (b), 
    and without pooling nor padding in time (c). 
    For more details see \cite{sercu2016advances}.
    *New results.}
\end{table}

Our implementation was done in Torch \cite{collobert11}.
We adopt the balanced sampling from \cite{sercu2015very}, by sampling
from context dependent state $CD_i$ with probability $p_i = \frac{f_i^\gamma}{\sum_j f_j^\gamma}$.
We keep $\gamma=0.8$ throughout the experiments during cross-entropy training.
During CE training, we optimize with simple SGD or NAG,
during ST we found NAG to be superior to SGD.
We regularize the stochastic sequence training by adding the gradient of 
cross-entropy loss, as proposed in \cite{su2013error}.

\subsection{Bidirectional LSTMs}

Given the recent popularity of LSTMs for acoustic
modeling~\cite{hannun14, sak15, miao15, mohamed15}, we have
experimented with such models on the Switchboard task using the Torch toolkit~\cite{collobert11}. We have looked
at the effect of the input features on LSTM performance, the number of
layers and whether start states for the recurrent layers should be
reset or carried over. We use bidirectional LSTMs that are trained
on non-overlapping subsequences of 20 frames. The subsequences coming from the same
utterance are contiguous so that the left-to-right final states for the current
subsequence can be copied to the left-to-right start states for the next
subsequence (i.e. carried over). For processing speed and in order to
get good gradient estimates, we group subsequences from multiple
utterances into minibatches of size 256. Regardless of the number of
LSTM layers, all models use a linear bottleneck of size 256 before the
softmax output layer (of size 32000).

In one experiment, we compare the effect of input features on
model performance. The baseline models are trained on 40-dimensional
FMLLR + 100-dimensional ivector frames and have 1024 (or 512) LSTM units per
layer and per direction (left-to-right and right-to-left). The forward
and backward activations from the previous LSTM layer
are concatenated and fed into the next LSTM layer.  The contrast model
is a single layer bidirectional LSTM trained on 128-dim features obtained
by performing PCA on 512-dimensional bottleneck features. The features are obtained from a 6-layer
DNN cross entropy trained on blocks of 11 consecutive FMLLR frames and 100-dimensional
i-vectors. In Table~\ref{lstm1}, we report recognition results on
Hub5'00 for these four models trained with 15 passes of cross-entropy SGD on the 300
hour (SWB-1) subset.

\begin{table}[htpb!]
\begin{center}
\begin{tabular}{|l|c|c|} \hline
     Model               & WER SWB & WER CH      \\ \hline
1-layer 1024 bottleneck  & 11.8 & 19.3 \\ \hline
2-layer 1024 FMLLR+ivec  & 11.1 & 19.2 \\ \hline
3-layer 1024 FMLLR+ivec  & 11.0 & 18.5 \\ \hline
4-layer 512 FMLLR+ivec   & 10.8 & 19.3 \\ \hline
\end{tabular}
\end{center}
\caption{\label{lstm1}
Word error rates on Hub5 2000 for various LSTM models trained with cross-entropy on 300 hours.}
\end{table}

\begin{figure}[htpb!]
\centerline{\includegraphics[scale=0.7]{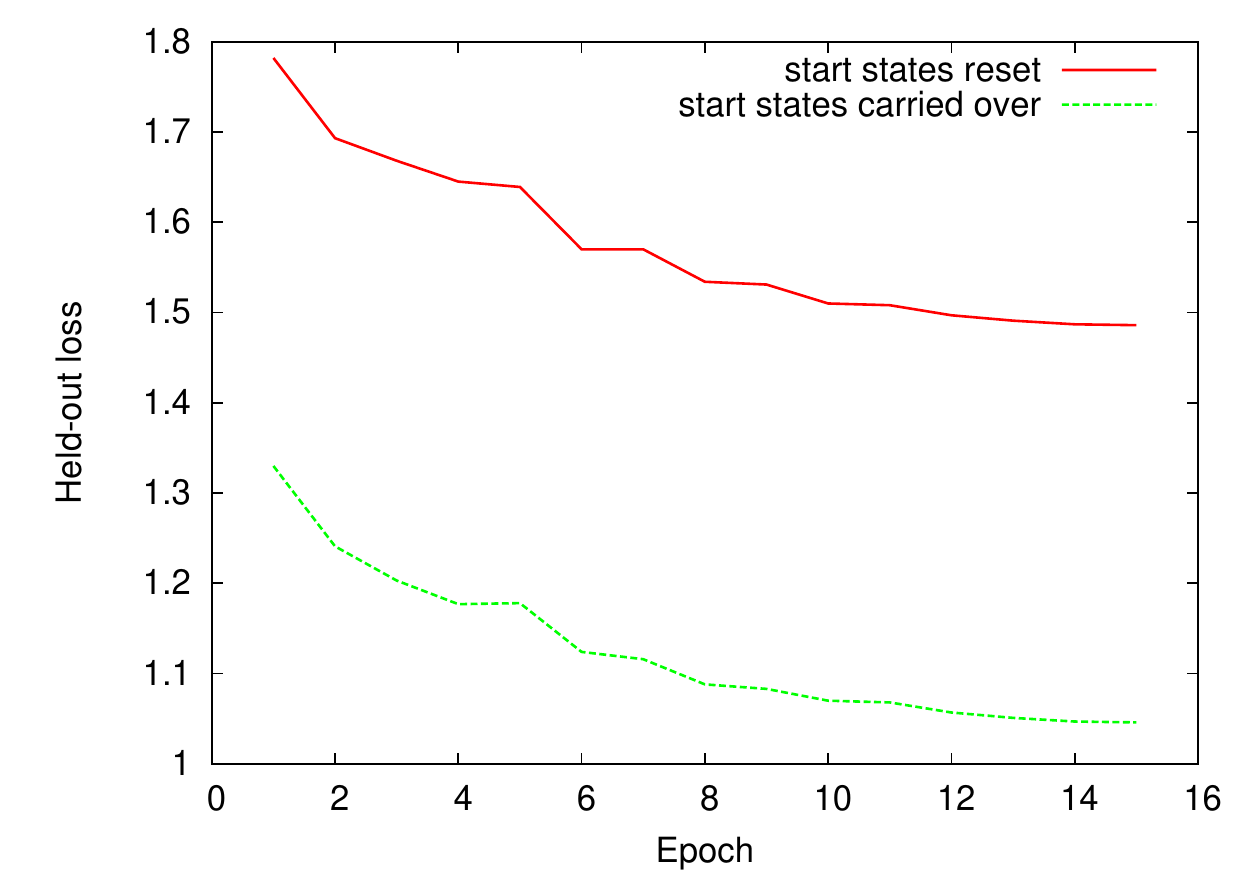}}
\caption{\label{fig2}  Single layer LSTM cross-entropy loss on held-out data with left-to-right start states which are either reset or carried over.}
\end{figure}

Due to a bug that affected our earlier multi-layer LSTM results, we
decided to go ahead with single layer bidirectional LSTMs on
bottleneck features on the full 2000 hour training set. We also
experimented with how to deal with the start states at the beginning
of the left-to-right pass. One option is to carry them over from the
previous subsequence and the other one is to reset the start states at
the beginning of each subsequence.  In Figure~\ref{fig2} we compare
the cross-entropy loss on held-out data between these two models.

As can be seen, the LSTM model with carried over start states is much
better at predicting the correct HMM state. However, when comparing
word error rates in Table~\ref{lstm2}, the LSTM with start states that
are reset has a better performance. We surmise that this is because
the increased memory of the LSTM with carried over start states is in
conflict with the state sequence constraints imposed by the HMM
topology and the language model.  Additionally, we show the WERs of
the DNN used for the bottleneck features and of a 4-layer 512 unit
LSTM. We observe that the 4 layer LSTM is significantly better than
the DNN and the two single layer LSTMs trained on bottleneck features.

\begin{table}[htpb!]
\begin{center}
\begin{tabular}{|l|c|c|c|c|} \hline
    Model & \multicolumn{2}{|c|}{WER SWB} & \multicolumn{2}{|c|}{WER CH} \\ \cline{2-5}
                        & CE   & ST   & CE   & ST     \\ \hline
6-layer DNN             & 11.7 & 10.3 & 18.5 & 17.0   \\ \hline
1-layer LSTM (carry-over) & 10.9 & --   & 18.3 & --     \\ \hline
1-layer LSTM (reset)      & 10.5 & 10.0 & 17.6 & 16.8   \\ \hline
4-layer LSTM (reset)      & 9.5  & 9.0  & 15.7 & 15.1   \\ \hline
\end{tabular}
\end{center}
\caption{\label{lstm2}
Word error rates on Hub5 2000 for DNN and LSTM models. All models are trained on 2000 hours with cross-entropy and sequence discriminative training.}
\end{table}

\subsection{Model combination}
In Table~\ref{comb} we report the performance of the individual models (RNN, VGG and 4-layer LSTM)
described in the previous subsections as well as the results after
frame-level score fusion. All decodings are done with a 30K word
vocabulary and a small 4-gram language model with 4M n-grams. We note
that RNNs and VGG nets exhibit similar performance and have a strong
complementarity which improves the WER by 0.6\% and 0.9\% on SWB and
CH, respectively.

\begin{table}[htpb!]
\begin{center}
\begin{tabular}{|l|c|c|} \hline
     Model   & WER SWB & WER CH      \\ \hline
RNN maxout   & 9.3 & 15.4        \\ \hline
VGG (a) from Table 2 & 9.4 & 15.7 \\ \hline
LSTM (4-layer, 512) & 9.0 & 15.1 \\ \hline
RNN+VGG      & 8.7 & 14.5 \\ \hline
VGG+LSTM     & 8.6 & 14.6 \\ \hline
RNN+VGG+LSTM & 8.6 & 14.4 \\ \hline
\end{tabular}
\end{center}
\caption{\label{comb}
Word error rates for individual acoustic models and frame-level score fusions on Hub5 2000.}
\end{table}

\subsection{Language modeling experiments}
\label{LM}

Our language modeling strategy largely parallels that described
in~\cite{saon15}.  For completeness, we will repeat some of the
details here.  The main difference is an increase in the vocabulary
size from 30K words to 85K words.

When comparing acoustic models in previous sections, we used a
relatively small legacy language model used in previous publications:
a 4M n-gram (n=4) language model with a vocabulary of 30.5K words.
We wanted to increase the language model coverage in a manner that
others can replicate.  To this end, we increased the vocabulary size
from 30.5K words to 85K words by adding the vocabulary of the publicly
available Broadcast News task.  We also added to the LM publicly
available text data from LDC, including Switchboard, Fisher, Gigaword,
and Broadcast News and Conversations.  The most relevant data are the
transcripts of the 1975 hour audio data used to train the acoustic
model, consisting of about 24M words.

For each corpus we trained a 4-gram model with modified Kneser-Ney
smoothing~\cite{chen99}.  The component LMs are linearly interpolated
with weights chosen to optimize perplexity on a held-out set.  Entropy
pruning~\cite{stolcke98} was applied, resulting in a single 4-gram LM
consisting of 36M n-grams.  This new n-gram LM was used together with
our best acoustic model to decode and generate word lattices for LM
rescoring experiments.  The first two lines of Table~\ref{lm-tab} show
the improvement using this larger n-gram LM with larger vocabulary
trained on more data.  The WER improved by 1.0\% for SWB.  Part of
this improvement (0.1-0.2\%) was due to also using a larger beam for
decoding and a change in vocabulary tokenization.

\begin{table}[htpb!]
\begin{center}
\begin{tabular}{|l|c|c|} \hline
LM                                 & WER SWB & WER CH \\ \hline 
30K vocab, 4M n-grams              & 8.6     & 14.4   \\ \hline 
85K vocab, 36M n-grams             & 7.6     & 13.7   \\ \hline
n-gram + model M                   & 7.0     & 12.6   \\ \hline
n-gram + NNLM                      & 6.8     & 12.4   \\ \hline
n-gram + model M + NNLM            & 6.6     & 12.2   \\ \hline 
\end{tabular}
\end{center}
\caption{\label{lm-tab}
Comparison of word error rates for different LMs.}
\end{table}

We used two types of LMs for LM rescoring: model M, a class-based
exponential model~\cite{chen09} and feed-forward neural network LM
(NNLM)~\cite{Bengio03,Emami06,Schwenk07,emamiasru07}.  We built a
model M LM on each corpus and interpolated the models, together with
the 36M n-gram LM.  As shown in Table~\ref{lm-tab}, using model M
results in an improvement of 0.6\% on SWB.

We built two NNLMs for interpolation. One was trained on just the most
relevant data: the 24M word corpus (Switchboard/Fisher/CallHome
acoustic transcripts).  Another was trained on a 560M word subset of
the LM training data: in order to speed up training for this larger
set, we employed a hierarchical NNLM
approximation~\cite{Emami06,kuo2012large}.  Table~\ref{lm-tab} shows
that the NNLMs provided an additional 0.4\% improvement over the model
M result on SWB. Compared with the n-gram LM baseline, LM rescoring
yielded a total improvement of 1.0\% on SWB (7.6\% to 6.6\%) and 1.5\%
on CH (13.7\% to 12.2\%).

\section{Conclusion}
\label{conclusion}
In our previous Switchboard system paper~\cite{saon15} we have
observed a good complementarity between recurrent nets and
convolutional nets and their combination led to significant accuracy
gains. In this paper we have presented an improved unfolded RNN (with
maxout instead of sigmoid activations) and a stronger CNN obtained by
adding more convolutional layers with smaller kernels and ReLu
nonlinearities. These improved models still have good complementarity
and their frame-level score combination in conjunction with a
multi-layer LSTM leads to a 0.4\%-0.7\% decrease in WER over the LSTM.
Multi-layer LSTMs were the strongest performing model followed closely
by the RNN and VGG nets. We also believe that LSTMs have more
potential for direct sequence-to-sequence modeling and we are actively
exploring this area of research. On the language modeling side, we
have increased our vocabulary from 30K to 85K words and updated our
component LMs.

At the moment, we are less than 3\% away from achieving human performance
on the Switchboard data (estimated to be around 4\%). Unfortunately,
it looks like future improvements on this task will be considerably
harder to get and will probably require a breakthrough in direct
sequence-to-sequence modeling and a significant increase in training
data.
\section{Acknowledgment} 
The authors wish to thank E. Marcheret, J. Cui and M. Nussbaum-Thom for useful suggestions about LSTMs.

\bibliographystyle{IEEEtran}
\bibliography{inter2016}

\end{document}